\documentclass[10pt,twocolumn,letterpaper]{article}

\usepackage[pagenumbers]{cvpr}

\usepackage[hyphens]{url}
\usepackage{graphicx}
\usepackage{booktabs}
\usepackage{multirow}
\usepackage{amsmath}
\usepackage{amssymb}
\usepackage{caption}
\usepackage{float}

\definecolor{cvprblue}{rgb}{0.21,0.49,0.74}
\usepackage[pagebackref,breaklinks,colorlinks,allcolors=cvprblue]{hyperref}

\newcommand{\hut}[1]{\textcolor{black}{#1}}
\title{Efficient Audio-Visual Generation via Synchrony-Aware \\
Cross-Modal Sparse Attention}

\author{
Shengchuan Gao$^{1*}$ \quad
Teng Hu$^{1*}$ \quad
Bohao Feng$^{2*}$ \quad
Luchen Li$^1$\\
Wenqiang Wang$^3$ \quad
Hongqian Deng$^3$ \quad
Ran Yi$^{1\dagger}$\\[3pt]
$^1$Shanghai Jiao Tong University \quad
$^2$Alibaba Token Hub, Alibaba Group \quad
$^3$Alibaba Cloud Computing\\
{\tt\small \{gscdy111, hu-teng, 66orange, ranyi\}@sjtu.edu.cn}\\
{\tt\small \{fengbohao.fbh, channing.wwq\}@alibaba-inc.com \quad hongqiandeng@foxmail.com}\\
{\small $^*$Equal contribution.
\quad $^\dagger$Corresponding author.}\\[5pt]
{\small\texttt{Project page: }
\href{https://sjtugsc.github.io/synchrony-aware-av-generation/}
{\nolinkurl{https://sjtugsc.github.io/synchrony-aware-av-generation/}}}
}

\begin{document}

\maketitle

\begin{abstract}
Recent audio-visual generation models can synthesize synchronized video and sound in a unified diffusion process, but their inference cost remains high because long video token sequences require repeated attention computation across denoising steps.
A variety of acceleration techniques have been developed for video generation models, including low-bit quantization, attention sparsification, and feature caching.
However, since these methods are originally designed for video generation, directly applying them to audio-visual models overlooks the interactions between the audio and video branches and may therefore disrupt audio-video synchronization.
We present a synchronization-aware acceleration framework for efficient audio-visual generation.
Our key observation is that bidirectional audio-video cross-attention reveals structured interactions between the two branches, with high responses often concentrated on a few sound-related visual and temporal regions.
Guided by this interaction pattern, we introduce a protected sparse attention strategy that preserves high-fidelity computation for synchronization-critical tokens while sparsifying redundant attention interactions.
By explicitly accounting for cross-modal dependence during acceleration, our method improves inference efficiency while  keeping video quality, audio quality, and audio-video synchronization.
\end{abstract}

\section{Introduction}
Audio-visual generation~\cite{low2025ovi,hacohen2026ltx,hu2025harmony,zhang2026uniavgen,hu2026evolution,chen2026omni,chen2026cinedance} is moving from isolated video or audio synthesis~\cite{wan2025wan,kong2024hunyuanvideo,hu2025hunyuancustom,liang2026goku,wang2026poseanything,hu2026ultragen,xue2025ultravideo} toward unified models that jointly generate what is seen and what is heard.
Recent systems such as Ovi~\cite{low2025ovi}, Harmony~\cite{hu2025harmony}, and LTX-2~\cite{hacohen2026ltx} couple video and audio diffusion backbones through bidirectional cross-modal interaction, enabling generated speech, ambient sound, and event sounds to follow the visual scene.
This shift makes audio-video synchronization a first-order generation objective rather than a post-processing problem.
At the same time, unified audio-visual generation inherits the heavy inference cost of modern video diffusion transformers: dense attention over long spatio-temporal token sequences must be repeatedly evaluated over many denoising steps.

\begin{figure}[t]
\centering
\includegraphics[width=0.95\linewidth]{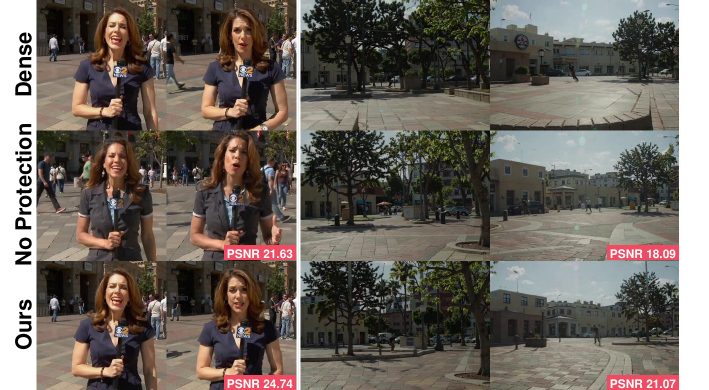}
\caption{
\textbf{Our method accelerates audio-visual generation while preserving high visual fidelity.}
On Ovi (left) and LTX-2.3 (right), our synchronization-aware protection better preserves frame quality and remains closer to dense inference , while achieving nearly $2\times$ and $1.4\times$ end-to-end speedup.
}
\label{fig:teaser}
\vspace{-2mm}
\end{figure}

\hut{A growing body of work accelerates video diffusion inference through efficient attention kernels, low-bit attention computation, sparse attention patterns, and intermediate-feature caching~\cite{flashattention2022,flashattention22023,sageattention2024,sageattention22024,sparsevideogen2025,teacache2024,fastercache2024,bwcache2025}.
These approaches exploit the substantial spatial and temporal redundancy present in video diffusion models.
However, directly transferring them to unified audio-visual generation introduces an additional challenge.
Existing acceleration methods generally determine which computation to sparsify, approximate, or reuse according to redundancy measured within an individual branch.
In an audio-visual generator~\cite{hu2025harmony,team2026mova}, however, a token or feature that appears redundant within one modality may still provide important evidence to the other modality through bidirectional cross-modal interaction.
Consequently, modality-agnostic acceleration may remove or reuse synchronization-critical computation, weakening the correspondence between generated sounds and visual events.}


\hut{Therefore, accelerating audio-visual generation requires more than identifying redundant computation.
The central challenge is to distinguish computation that can be safely approximated from computation that carries important cross-modal evidence.
Our key premise is that this distinction cannot be determined from either the video or audio stream independently.
Instead, acceleration decisions should explicitly account for the mutual dependencies established between the two modalities during joint denoising.}


\hut{To understand which computation is important to cross-modal synchronization, we analyze the bidirectional attention between the video and audio branches.
We observe that cross-modal interaction is highly structured and non-uniform.
In the video-to-audio direction, audio queries concentrate on a sparse subset of video tokens associated with sound-producing objects, temporally aligned events, and other synchronization-relevant visual content.
Conversely, in the audio-to-video direction, video queries exhibit selective dependence on audio tokens that characterize speech, acoustic events, and temporal sound cues.
Moreover, these salient interaction patterns remain relatively stable across nearby denoising steps.
These observations indicate that the model's native bidirectional cross-modal attention provides an intrinsic and reusable importance signal for identifying synchronization-critical computation.}


\hut{Based on these observations, we propose a training-free audio-visual interaction-aware acceleration framework that uses \textbf{bidirectional cross-modal saliency} as a shared guidance signal for both audio and video branches. We first aggregate the two directions of audio-video attention into modality-specific saliency maps, which characterize the importance of video tokens to the audio branch and the importance of audio tokens to the video branch, respectively. Since the current cross-modal attention is generally evaluated after modality-specific self-attention within the same block, we introduce \textbf{lagged cross-modal saliency reuse} to propagate the most recent interaction patterns across nearby denoising steps without an additional profiling pass. Guided by the cached saliency maps, we then propose \textbf{synchronization-aware protected sparse attention}, which preserves dense attention computation for synchronization-critical tokens in both the video and audio branches while sparsifying low-saliency interactions. We further introduce a \textbf{synchronization-aware cache-reuse mechanism} that jointly considers modality-specific feature variations, changes in salient regions, and the drift of bidirectional cross-modal dependencies, thereby preventing unsafe reuse that may disrupt audio-video alignment. Together, these components directly address the efficiency--quality--synchronization trade-off in unified audio-visual generation.}

Our contributions are summarized as follows:
\begin{itemize}
    \item We analyze \textbf{cross-modal interactions} in dual-branch audio-visual generation models and observe that bidirectional audio-video attention highlights sound-related visual and temporal regions, providing an internal saliency signal for synchronization-critical content.

\item \hut{We propose a \textbf{training-free synchronization-aware protected sparse attention mechanism}. Guided by modality-specific saliency maps, it preserves high-fidelity attention computation for synchronization-critical tokens in both audio and video branches, while sparsifying low-saliency interactions to reduce redundant computation.} 

\item \hut{We introduce a \textbf{synchronization-aware cache-reuse mechanism} that jointly considers modality-specific feature variations, changes in cross-modally salient tokens, and the drift of bidirectional audio-video dependencies. The proposed criterion can be integrated into general feature-caching methods, and we instantiate it with block-level caching to prevent unsafe reuse that may disrupt audio-video synchronization.}



    \item 
    We validate our method on open-source audio-visual generators, achieving efficient inference while preserving video fidelity, audio quality, and synchronization.
\end{itemize}
\begin{figure*}[t]
\centering
\includegraphics[width=0.95\textwidth]{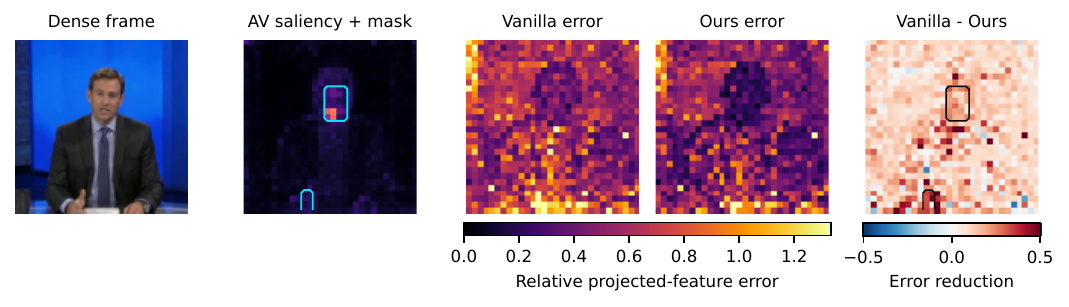}
\caption{
Motivation of synchronization-aware protection.
From left to right: a generated video frame, the extracted audio-visual saliency and protected regions, feature error of vanilla sparse attention, feature error of our protected sparse attention, and their error difference.
Red areas in the difference map indicate regions where our method reduces the approximation error compared with vanilla sparsification.
}
\label{fig:motivation}
\end{figure*}
\section{Related Work}

\subsection{Joint Audio-Visual Generation}

Audio-visual generation has been studied through video-to-audio synthesis, audio-conditioned visual generation, and joint audio-video diffusion.
Video-to-audio systems such as FoleyCrafter~\cite{foleycrafter2024}, MMaudio~\cite{cheng2025mmaudio}, and Hunyuan-Foely~\cite{shan2025hunyuanvideo} synthesize synchronized sound from visual content by aligning audio generation with video semantics and timing.
Other works connect pretrained visual and audio diffusion models through shared latent or cross-modal conditioning spaces~\cite{seeinghearing2024,avlink2024}.
More recent joint generators, including Ovi~\cite{low2025ovi}, Harmony~\cite{hu2025harmony}, Jova~\cite{huang2025jova}, and LTX-2~\cite{hacohen2026ltx}, move toward unified dual-stream architectures where video and audio branches exchange information through bidirectional attention.
These models make cross-modal interaction central to generation quality: synchronization emerges from the repeated exchange of semantic and temporal information rather than from a separate post-processing module.

Our work is built on this trend, but studies a different question.
Rather than improving the generative backbone itself, we ask how to accelerate such backbones without weakening the cross-modal evidence used for audio-video synchronization.
This is different from ordinary video generation acceleration because the importance of a token or block cannot be determined only from visual redundancy.

\subsection{Efficient Attention for Video Diffusion Models}

Attention is a major bottleneck in  video diffusion transformers.
System-level methods such as FlashAttention~\cite{flashattention2022,flashattention22023,shah2024flashattention} accelerate exact attention by reducing memory traffic and improving GPU work partitioning, while low-bit methods such as SageAttention~\cite{sageattention2024} and SageAttention2~\cite{sageattention22024} use quantized computation.
These methods preserve dense attention and are complementary to sparsification.

Another line of work reduces redundant attention interactions.
Sparse VideoGen~\cite{sparsevideogen2025,yang2026sparse,fujitake2022video} exploits spatial and temporal sparsity in video diffusion attention, while related sparse or block-sparse methods~\cite{wu2025vmoba,yang2026sparse,fujitake2022video} select important token blocks or adapt sparsity to video structure.
Cache-based methods\cite{kahatapitiya2025adaptive} instead reuse intermediate representations across denoising steps.
TeaCache~\cite{teacache2024}, FasterCache~\cite{fastercache2024}, BWCache~\cite{bwcache2025}, and SeaCache\cite{chung2026seacache} adopt different reuse criteria based on feature changes, timestep behavior, block redundancy, or CFG redundancy.
Their effectiveness stems from the substantial spatial, temporal, and denoising-step redundancy in video diffusion models.

However, existing acceleration methods are mainly designed for video-only generation, with sparse or cache decisions based on intra-modal redundancy, such as video attention patterns or feature similarity across timesteps.
Direct application to audio-visual generation can be risky, because a visually redundant token or stable block may still provide important evidence for the audio branch and audio-video synchronization.
In contrast, our method uses cross-modal saliency to identify and protect synchronization-critical computation during sparse attention and cache reuse.

\section{Motivation and Observation}
\label{sec:motivation}

\paragraph{Preliminaries: Dual-stream architecture and bottleneck.}
Different from video-only diffusion models, joint audio-visual generators maintain separate video and audio streams and exchange information through cross-modal attention.
A typical Ovi-style block contains modality-specific self-attention and FFN layers, where text attention provides conditioning and bidirectional audio-video attention enables cross-modal interaction.
This design makes synchronization a result of repeated interaction between two branches during denoising.
However, it also changes the acceleration problem: the number of video tokens is much larger than the number of audio tokens, so video self-attention and video-related attention modules dominate the latency.
Meanwhile, computations that appear redundant within the video branch may provide important evidence for the audio branch.
Therefore, directly applying video-oriented sparse attention or cache reuse may skip synchronization-critical computation.


\paragraph{Cross-modal interaction as acceleration guidance.}
To identify which computation should be protected, we examine the cross-modal attention patterns of dual-stream audio-visual generators during inference.
We observe that bidirectional audio-video attention is highly non-uniform: only a small subset of spatiotemporal video regions and audio tokens receives strong responses from the opposite modality.
These regions may be inconspicuous under intra-modal redundancy measures, but often carry information important for audio generation and audio-video synchronization.
As illustrated in Fig.~\ref{fig:motivation}, the saliency-derived protected regions coincide with locations where vanilla sparse attention introduces larger feature errors, while our protected computation reduces such errors.
The same principle also applies to the audio branch through reverse cross-modal saliency.
Therefore, cross-modal attention can serve as a model-internal saliency signal for synchronization-aware acceleration, guiding the model to protect high-saliency computation and apply sparse computation or feature reuse to low-saliency regions.

Fig.~\ref{fig:motivation} provides an empirical illustration of these observations.
These results suggest cross-modal interaction can serve as a practical signal for synchronization-aware acceleration.

\section{Method}

\begin{figure*}[t]
\centering
\includegraphics[width=0.96\textwidth]{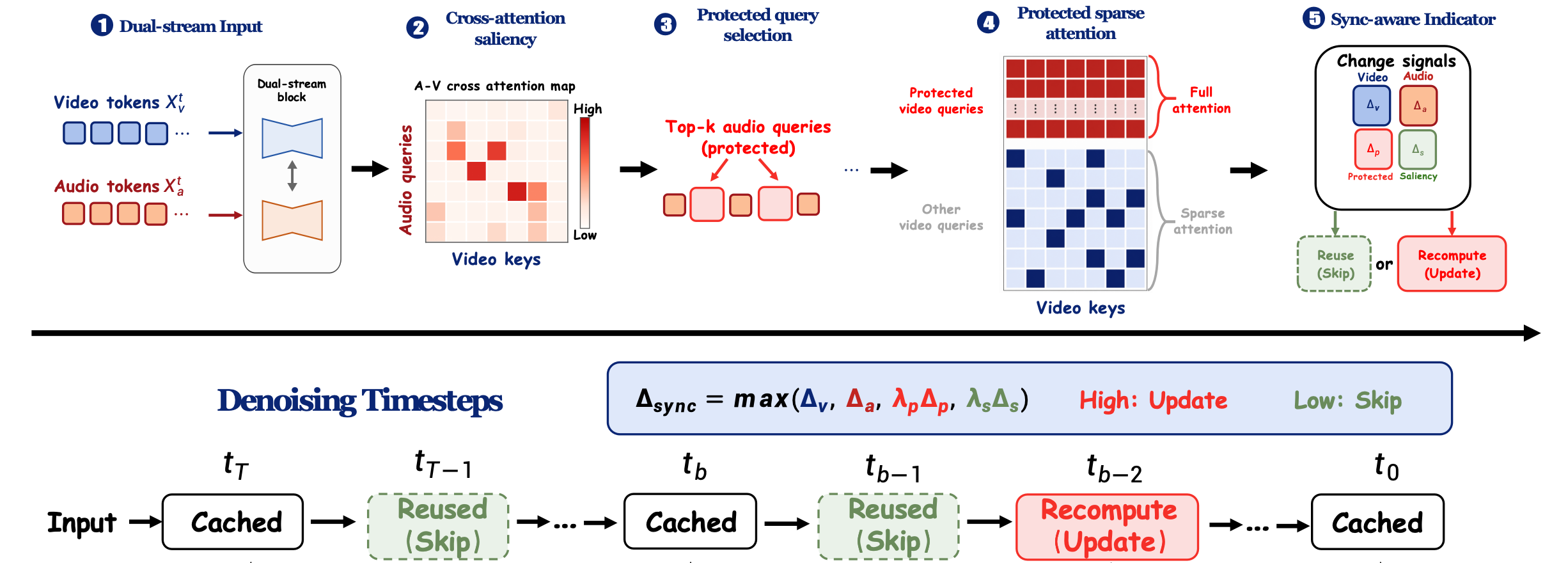}
\caption{
Overview of the proposed synchronization-aware acceleration framework.
Cross-modal interactions between audio and video branches are used to guide protected sparse attention and cache reuse.
}
\label{fig:method_overview}
\end{figure*}

\paragraph{Framework overview.}
\hut{The observations in Sec.~\ref{sec:motivation} suggest that acceleration for audio-visual generation should distinguish generic computational redundancy from synchronization-critical computation.
Based on this principle, we propose a synchronization-aware acceleration framework that uses bidirectional cross-modal interaction as a shared importance prior at two complementary granularities.
At token level, we identify synchronization-critical tokens in both the video and audio branches and preserve high-fidelity self-attention computation for these tokens.
At block level, we incorporate modality-specific feature changes and bidirectional cross-modal saliency into the cache-reuse criterion, preventing unsafe reuse when synchronization-critical content or audio-video dependencies change substantially.}

\hut{A practical challenge is the cross-modal attention used to estimate importance is evaluated after modality-specific self-attention within a dual-stream block.
The current cross-modal saliency therefore cannot guide self-attention that has  been executed at the same denoising step.
We address this issue with lagged saliency reuse, exploiting the empirical stability of cross-modal interaction across neighboring denoising steps.
As illustrated in Fig.~\ref{fig:method_overview}, the resulting framework consists of three components:
bidirectional cross-modal saliency estimation, audio-guided protected sparse attention, and synchronization-aware block reuse.}

\subsection{Cross-Modal Saliency and Lagged Reuse}

\noindent\hut{\textbf{Bidirectional cross-modal saliency.}} Let $X_v \in \mathbb{R}^{N_v \times d}$ and $X_a \in \mathbb{R}^{N_a \times d}$ denote video and audio tokens in an audio-visual diffusion transformer block.
We first estimate synchronization-critical regions from bidirectional audio-video cross-attention.
For clarity, we describe the audio-to-video direction, where audio queries attend to video keys:
\begin{equation}
    A^{a \rightarrow v}_{t,\ell,h}
    =
    \mathrm{softmax}
    \left(
    Q^a_{t,\ell,h}(K^v_{t,\ell,h})^\top / \sqrt{d_h}
    \right),
\end{equation}
where $t$, $\ell$, and $h$ denote denoising step, layer, and attention head.
Aggregating attention mass over audio queries, heads, and selected layers gives a video-token saliency map:
\begin{equation}
    s^v_{t,i}
    =
    \mathrm{Norm}
    \left(
    \sum_{\ell \in \mathcal{L}_c}
    \sum_h
    \sum_{j=1}^{N_a}
    A^{a \rightarrow v}_{t,\ell,h}[j,i]
    \right).
\end{equation}
The reverse video-to-audio direction is handled analogously to obtain audio-token saliency, reflecting the bidirectional dependence between the two branches.
Given the saliency score, we select a protected query set:
\begin{equation}
    \mathcal{P}_v=\mathrm{TopK}(\{s^v_{t,i}\}_{i=1}^{N_v}, \rho_v N_v),
\end{equation}
where $\rho_v$ is a small protection ratio.
The selected set identifies video queries that are likely to carry important cross-modal evidence for audio generation and synchronization.

\noindent\hut{\textbf{Lagged saliency reuse.}} A practical scheduling issue arises in Ovi-style fusion blocks: video self-attention is executed before later audio-video cross-attention in the same block.
Therefore, the current cross-attention map cannot directly guide the already executed video self-attention.
Instead of introducing an additional dense profiling pass, we use a lagged saliency cache.
At denoising step $t$, cross-modal attention updates a cached saliency map, and nearby future steps use it to select protected tokens:
\begin{equation}
    \bar{s}^v_t =
    \alpha \bar{s}^v_{t-1} + (1-\alpha)s^v_t .
\end{equation}
The cache is updated every $K$ denoising steps to reduce overhead and temporal flicker.
This design relies on the observation that high-saliency regions change smoothly across neighboring denoising steps.
Thus, the same cross-modal saliency signal can be reused in a local temporal window to guide both sparse attention protection and cache reuse.

\subsection{\hut{Synchronization-Aware Protected Sparse Attention}}

\hut{After obtaining the cached cross-modal saliency maps, we use them to protect synchronization-critical computation in sparse self-attention.
For modality $m\in{v,a}$, standard dense self-attention computes:}
\begin{equation}
\hut{
O^m_{\mathrm{dense}}
=
\mathrm{softmax}
\left(
\frac{Q^m(K^m)^\top}{\sqrt{d_h}}
\right)V^m,
}
\end{equation}
\hut{whose score-computation cost grows quadratically with the modality token length $N_m$.}

\noindent\hut{\textbf{Sparse attention.}}
\hut{A standard sparse-attention method restricts each query to a subset of keys specified by a sparse connectivity pattern
$\mathcal{S}*m$.
Equivalently, we define an additive attention mask:}
\begin{equation}
\hut{
M^m_{ij}
=
\begin{cases}
0, & (i,j)\in\mathcal{S}_m,\\
-\infty, & (i,j)\notin\mathcal{S}*m,
\end{cases}
}
\end{equation}
\hut{and compute:}
\begin{equation}
\hut{
O^m_{\mathrm{sparse}}
=
\mathrm{softmax}
\left(
\frac{Q^m(K^m)^\top}{\sqrt{d_h}}
+
M^m
\right)V^m.
}
\end{equation}
\hut{By masking a large fraction of query-key connections, sparse attention reduces redundant computation.
However, the sparse pattern is typically determined from redundancy within an individual modality and does not account for whether a query carries important evidence for the opposite branch.}

\noindent\hut{\textbf{Cross-modal protection.}}
\hut{At denoising step $t$, we use the cached modality-specific saliency maps to select protected query sets:}
\begin{equation}
    \mathcal{P}^{m}_t
    =
    \mathrm{TopK}
    \left(
        \left\{\bar{s}^{m}_{t-1,i}\right\}_{i=1}^{N_m},
        \left\lceil \rho_m N_m \right\rceil
    \right),
    \qquad m\in\{v,a\},
\end{equation}
\hut{where $\rho_m$ is the protection ratio for modality $m$.
We then modify the original sparse mask by restoring full key connectivity for protected queries:}
\begin{equation}
(\widetilde{M}^{m})_{t,ij}
=
\begin{cases}
0,
& i\in\mathcal{P}^{m}_t
\ \text{or}\
(i,j)\in\mathcal{S}_m,\\
-\infty,
& \text{otherwise}.
\end{cases}
\end{equation}
\hut{The resulting protected sparse attention is}
\begin{equation}
O^m
=
\mathrm{softmax}
\left(
\frac{Q^m(K^m)^\top}{\sqrt{d_h}}
+
\widetilde{M}^{m}_t
\right)V^m,
\qquad m\in{v,a}.
\end{equation}

\hut{Equivalently, ordinary queries follow the original sparse pattern, whereas protected queries retain complete dense attention rows:}
\begin{equation}
\resizebox{\linewidth}{!}{$\displaystyle
O^m_i
=
\begin{cases}
\mathrm{DenseAttn}(q^m_i,K^m,V^m),
& i\in\mathcal{P}^{m}*t,\\
\mathrm{SparseAttn}*{\mathcal{S}_m}(q^m_i,K^m,V^m),
& i\notin\mathcal{P}^{m}_t,
\end{cases}
\qquad m\in{v,a}.
$}
\end{equation}

\hut{This query-row protection does not merely retain the selected token values.
Instead, it preserves high-fidelity representation updates for tokens that are important to the opposite modality.
Sound-related visual tokens can therefore aggregate complete visual context before interacting with the audio branch, while synchronization-relevant audio tokens retain complete acoustic context before interacting with the video branch.
Low-saliency queries in both modalities continue to follow the original sparse computation path.
As a result, the proposed mechanism reduces redundant attention computation while preserving synchronization-critical representations and maintaining reliable audio-video alignment.
}

\subsection{Synchronization-Aware Cache Reuse}

\hut{Protected sparse attention reduces redundant token interactions within each denoising step, but substantial redundancy also exists across neighboring steps, where intermediate audio and video features often evolve smoothly.
Feature-caching methods exploit this temporal redundancy by reusing previously computed representations when the current features remain sufficiently similar to those at an anchor step.
However, directly applying such modality-agnostic criteria to audio-visual generation can be unsafe.
A small global feature difference in either branch does not necessarily indicate that the underlying cross-modal dependency remains unchanged.
Localized variations, such as mouth motion, object contact, or a shift in the active sound source, may involve only a small subset of tokens while still being critical to audio-video synchronization.
Therefore, cache reuse should account not only for overall feature similarity, but also for changes in synchronization-critical regions and bidirectional cross-modal interactions.}

\hut{We therefore introduce a synchronization-aware cache-reuse criterion and instantiate it with block-level residual caching.}
Let $(Y^v_{t,\ell},Y^a_{t,\ell})$ be the video and audio outputs of block $\ell$ at step $t$, and let $(\hat{Y}^v_{\ell},\hat{Y}^a_{\ell})$ be the cached outputs from the previous anchor step.
The ordinary stream-wise feature changes are:
\begin{equation}
    \Delta^v_{t,\ell}
    =
    \frac{\|Y^v_{t,\ell}-\hat{Y}^v_{\ell}\|_1}
    {\|Y^v_{t,\ell}\|_1+\epsilon},
    \quad
    \Delta^a_{t,\ell}
    =
    \frac{\|Y^a_{t,\ell}-\hat{Y}^a_{\ell}\|_1}
    {\|Y^a_{t,\ell}\|_1+\epsilon}.
\end{equation}
To make the cache sensitive to synchronization-critical regions, we additionally compute protected-region changes:
\begin{equation}
    \Delta^{p}_{t,\ell}
    =
    \frac{\|Y^v_{t,\ell}[\mathcal{P}_v]-\hat{Y}^v_{\ell}[\mathcal{P}_v]\|_1}
    {\|Y^v_{t,\ell}[\mathcal{P}_v]\|_1+\epsilon}.
\end{equation}
We also measure saliency drift between the current and cached cross-modal saliency maps:
\begin{equation}
    \Delta^{s}_{t}
    =
    1 -
    \frac{\langle \bar{s}^v_t, \hat{s}^v \rangle}
    {\|\bar{s}^v_t\|_2\|\hat{s}^v\|_2+\epsilon}.
\end{equation}
The final synchronization-aware cache indicator is
\begin{equation}
    \Delta^{\mathrm{sync}}_{t,\ell}
    =
    \max
    \left(
    \Delta^v_{t,\ell},
    \Delta^a_{t,\ell},
    \lambda_p \Delta^p_{t,\ell},
    \lambda_s \Delta^s_t
    \right).
\end{equation}
A block output is reused only when $\Delta^{\mathrm{sync}}_{t,\ell} < \tau$,
where $\tau$ is a cache threshold.
Compared with original BlockCache, this criterion is audio-visual aware in three aspects:
it considers both video and audio streams, emphasizes protected sound-source regions, and prevents reuse when cross-modal saliency changes significantly.
In practice, we apply this cache only to middle denoising steps and selected middle blocks, where diffusion features are more stable and cache reuse is less likely to disturb generation quality.

\subsection{Inference Pipeline and Complexity}

The full inference pipeline contains three components.
First, selected audio-video cross-attention layers update the lagged saliency cache.
Second, video self-attention uses sparse attention for ordinary queries and dense-row protection for high-saliency queries.
Third, eligible fusion blocks use synchronization-aware cache reuse to skip redundant block computation when both stream-wise features and cross-modal saliency remain stable.

Let $\kappa$ denote the average sparse keep ratio of video self-attention.
Dense video attention has score complexity $O(HN_v^2d_h)$.
Our method has approximate complexity:
\begin{equation}
    O\left(H(\kappa N_v^2+\rho_v N_v^2)d_h\right),
\end{equation}
where the first term is sparse attention and the second term is dense-row protection.
Since $\rho_v$ is small, the protection overhead is limited.
The synchronization-aware cache further reduces computation by skipping reusable fusion blocks while avoiding reuse when audio-video interaction changes.
Overall, the method is training-free and can be applied to open-source dual-branch audio-visual generators with accessible cross-attention and block outputs.

\begin{table*}[t]
\centering
\small
\setlength{\tabcolsep}{5pt}
\caption{
Main comparison on Ovi and LTX-2.3.
}
\label{tab:main_results}
\resizebox{0.8\linewidth}{!}{%
\begin{tabular}{ll|cccc|ccc|cc}
\toprule
\multirow{2}{*}{Base Model} &
\multirow{2}{*}{Method} &
\multicolumn{4}{c|}{Video Quality} &
\multicolumn{3}{c|}{Audio / Sync.} &
\multicolumn{2}{c}{Efficiency} \\
\cmidrule(lr){3-6}
\cmidrule(lr){7-9}
\cmidrule(lr){10-11}
& &
PSNR$\uparrow$ & LPIPS$\downarrow$ & SSIM$\uparrow$ & Q-Align$\uparrow$ &
Sync-C$\uparrow$ & Sync-D$\downarrow$ & VISQOL$\uparrow$ &
Speedup$\uparrow$ & Latency (s)$\downarrow$ \\
\midrule

\multirow{7}{*}{Ovi}
& Dense Baseline        & $\infty$ & 0.0000 & 1.0000 & 0.7953 & 4.688 & 8.172 & Ref & 1.000x & 448.75 \\
& SVG                   & 21.7630  & 0.2511 & 0.7706 & 0.7601 & 3.658 & 8.729 & 3.3670 & 1.300x & 344.83 \\
& BWCache               & 21.9480  & 0.2444 & 0.7672 & 0.7560 & 3.822 & 8.635 & 3.5192 & 1.774x & 252.71 \\
& TeaCache              & 24.7775  & 0.1494 & 0.8365 & 0.7546 & 3.743 & 8.651 & 3.4610 & 1.811x & 247.83 \\
& SeaCache              & 22.1419  & 0.2389 & 0.7743 & 0.7218 & 3.901 & \textbf{8.311} & 3.3588 & \textbf{2.471x} & \textbf{181.47} \\
& FasterCache              & 22.3227  & 0.2408 & 0.7836 & 0.7489 & 3.846 & 8.591 & 3.4012 & 1.483x & 302.48 \\
& Ours                  & \textbf{25.6857} & \textbf{0.1398} & \textbf{0.8385} & \textbf{0.7822} & \textbf{4.606} & 8.378 & \textbf{3.6902} & 1.992x & 225.30 \\

\midrule

\multirow{7}{*}{LTX-2.3}
& Dense Baseline        & $\infty$ & 0.0000 & 1.0000 & 0.8124 & 5.304 & 7.3081 & Ref & 1.000x & 255.41  \\
& SVG                   & 18.7640  & 0.3961 & 0.6193 & 0.7639 & 4.722 & 7.8021 & 3.2760 & 1.041x & 245.35 \\
& BWCache               & 19.5013  & 0.3690 & 0.6224 & 0.8013 & 4.838 & 7.7142 & 3.4199 & 1.227x & 208.15 \\
& TeaCache              & 18.8912  & 0.3756 & 0.6162 & 0.7874 & 4.576 & 7.5110 & 3.3419 & 1.269x & 201.30 \\
& SeaCache              & 19.1386  & 0.3689 & 0.6179 & 0.8027 & 4.869 & 7.5328 & 3.4527 & \textbf{1.325x} & \textbf{192.74} \\
& FasterCache              & 18.9218  & 0.3847 & 0.6190 & 0.7826 & 4.731 & 7.6654 & 3.3602 & 1.132x & 225.63 \\
& Ours                  & \textbf{20.8737} & \textbf{0.3194} & \textbf{0.6737} & \textbf{0.8052} & \textbf{5.134} & \textbf{7.3964} & \textbf{3.7638} & 1.281x & 199.38 \\

\bottomrule
\end{tabular}}
\end{table*}
\section{Experiments}


\subsection{Experimental Settings}



\paragraph{Base models and baselines.}
We evaluate our method on two open-source audio-visual generation models, Ovi and LTX-2.3.
For Ovi, we generate 141-frame, 5-second videos at $960\times960$ resolution with 50 denoising steps.
For LTX-2.3, we use its two-stage pipeline to generate 10-second videos at $1280\times704$ resolution and 24 FPS, with 30 low-resolution denoising steps followed by 3 high-resolution refinement steps.

We compare against the dense model and representative training-free video acceleration methods, including the sparse attention method SparseVideoGen(SVG)\cite{sparsevideogen2025} and cache-based methods BWCache\cite{bwcache2025}, TeaCache\cite{teacache2024}, SeaCache\cite{chung2026seacache}, and FasterCache\cite{fastercache2024}.
Since these baselines are designed for video generation, we apply their sparsification or cache-reuse strategies to the video branch of audio-visual backbones under the default inference setting.

\paragraph{Implementation details.}
For Ovi, our acceleration method is applied throughout the 50-step denoising process.
For LTX-2.3, acceleration is applied only to the 30-step low-resolution stage, while the spatial upsampling and high-resolution refinement stage remain unchanged.
All experiments are conducted on NVIDIA A800 GPUs.

\paragraph{Evaluation metrics.}
Following prior video-generation acceleration studies, we evaluate fidelity and efficiency using standard reconstruction and latency metrics.
For video quality, we report PSNR, SSIM\cite{wang2004image}, and LPIPS\cite{zhang2018unreasonable} against dense baseline outputs, along with Q-Align\cite{wu2023q} for perceptual assessment.
For joint audio-visual generation, we evaluate the audio branch in terms of audio-video synchronization and similarity to dense audio outputs.
Specifically, Sync-C and Sync-D\cite{chung2016out} measure synchronization, while VISQOL\cite{hines2015visqol} assesses audio similarity.
For efficiency, we report end-to-end latency and speedup over dense inference under the same sampling configuration.

\subsection{Main Results}

\paragraph{Quantitative Comparison.}
Table~\ref{tab:main_results} reports the main comparison on Ovi and LTX-2.3.
Across both models, our method achieves the best overall quality among accelerated methods, with gains covering not only video fidelity but also audio similarity and audio-video synchronization.
This confirms the benefit of protecting cross-modal saliency regions rather than applying sparsification or cache reuse in a modality-agnostic manner.
On Ovi, Ours consistently preserves visual fidelity, perceptual quality, audio similarity, and synchronization better than other accelerated baselines, while delivering around $2\times$ end-to-end acceleration.
On LTX-2.3, our method shows a similar trend, achieving better reconstruction quality and stronger audio metrics under comparable acceleration.

\paragraph{Efficiency Analysis.}
The overhead of our method comes from saliency extraction and dense-row recomputation for protected queries.
This cost is limited because only a small fraction of high-saliency queries is protected, while most tokens follow the sparse attention path.
In addition, the saliency map is updated intermittently and reused across nearby steps, avoiding per-step profiling overhead.
The synchronization-aware cache compensates for this cost by reusing stable block outputs.
Thus, our method preserves synchronization-critical computation while retaining the efficiency benefits of sparsification and caching.

\begin{figure}[t]
\centering
\includegraphics[width=1.0\linewidth]{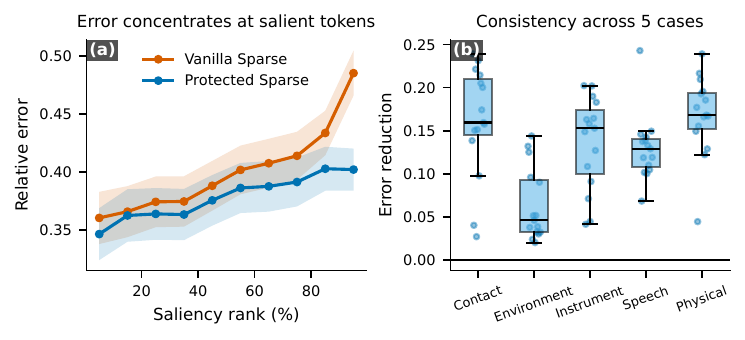}
\caption{
Saliency-conditioned feature error analysis.
(a) Feature error across different cross-modal saliency ranks.
(b) Error reduction on top-$5\%$ protected queries.
}
\label{fig:saliency_selective_preservation}
\vspace{-2mm}
\end{figure}

\begin{figure}[t]
\centering
\includegraphics[width=0.95\linewidth]{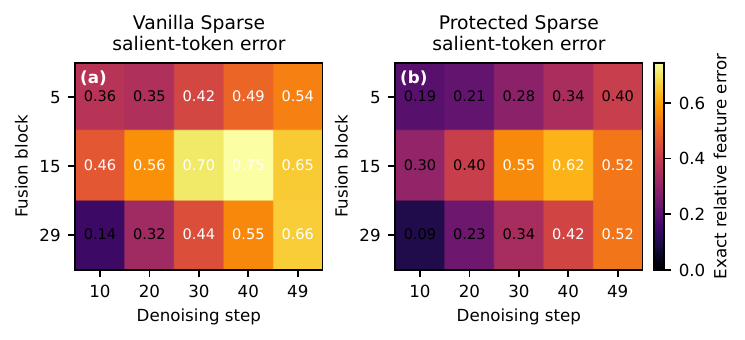}
\caption{
Layer--timestep comparison of feature error.
}
\label{fig:layer_timestep_exact_preservation}
\vspace{-2mm}
\end{figure}
\begin{figure*}[t]
\centering
\includegraphics[width=0.8\textwidth]{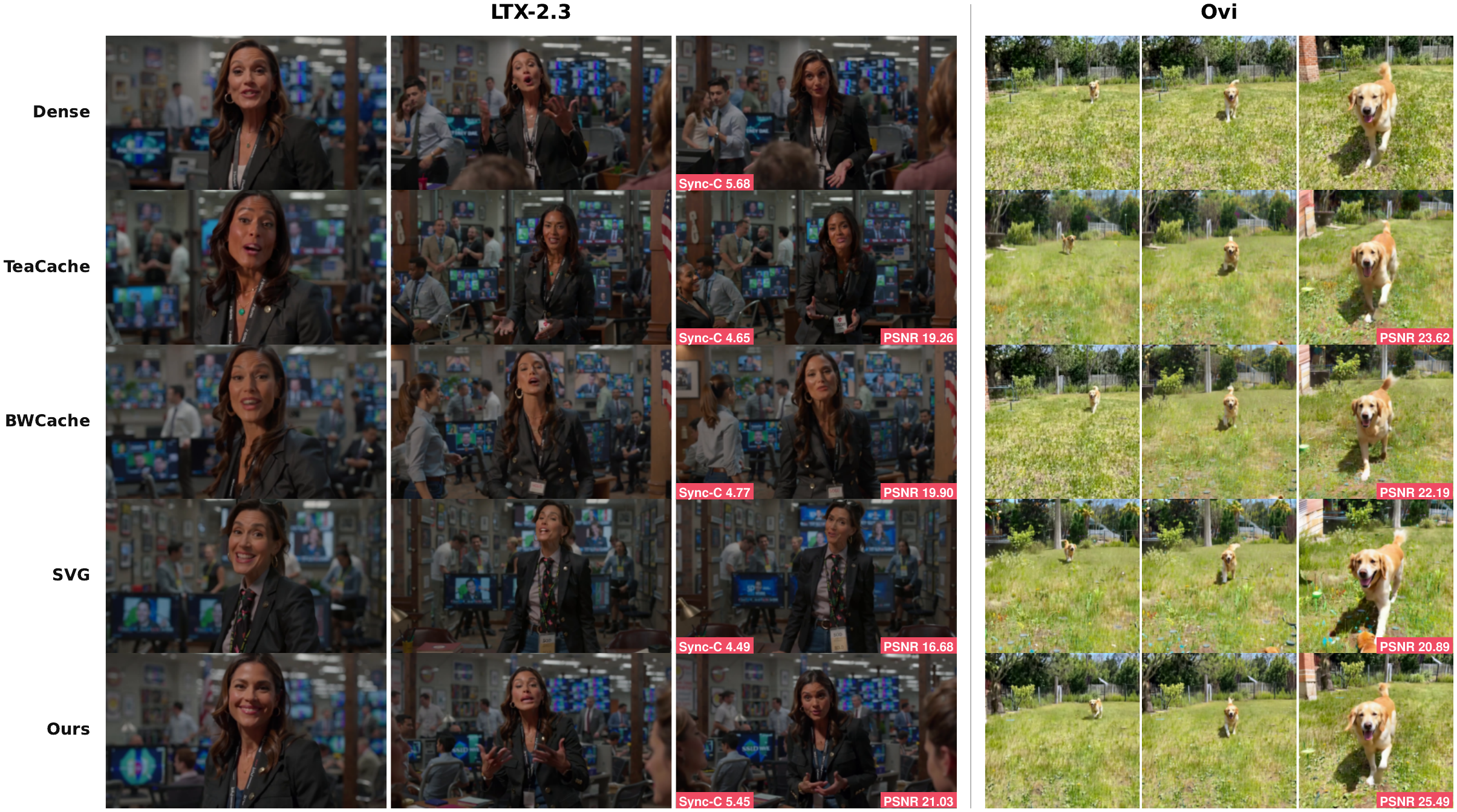}
\caption{
Qualitative comparison of dense inference, unprotected acceleration, and our protected acceleration.
The protected computation path better preserves salient visual structures and reduces local artifacts under acceleration.
}
\label{fig:qualitative_comparison}
\end{figure*}
\paragraph{Error Analysis.}
We further analyze the internal video self-attention error by comparing accelerated outputs with dense inference under the same prompt, seed, layer, and denoising step.
Fig.~\ref{fig:saliency_selective_preservation}(a) compares vanilla sparsification and our protected sparsification across tokens with different cross-modal saliency ranks.
The error of vanilla sparsification increases on more salient tokens, while our method clearly reduces the error in the most important token regions.
Although protection is applied only to selected high-saliency queries, we also observe moderate error reduction on less salient tokens, suggesting that preserving key cross-modal representations can benefit the surrounding generation trajectory.

Fig.~\ref{fig:saliency_selective_preservation}(b) summarizes the error reduction on the actual top-$5\%$ protected tokens across different semantic categories.
The reduction remains positive across representative cases, showing that the benefit is not limited to a single favorable example.
Fig.~\ref{fig:layer_timestep_exact_preservation} further examines the effect along model depth and denoising steps.
Compared with vanilla sparsification, protected sparsification yields lower feature error over the sampled fusion blocks and timesteps, especially in stages where sparse approximation strongly affects the subsequent trajectory.
These results indicate that our method selectively protects cross-modal salient queries that are most vulnerable to sparse approximation, while also improving the stability of nearby representations.

\subsection{Ablation Studies}

\paragraph{Component ablation.}
Table~\ref{tab:component_ablation} studies the contribution of each protection component under the same backbone.
Without protection, sparse computation causes clear degradation in both visual quality and synchronization.
Adding attention protection substantially improves the results, indicating that dense recomputation for cross-modal salient queries effectively reduces the error introduced by sparse attention.
Cache protection brings consistent gains by avoiding unsafe feature reuse, although its effect is smaller than attention protection.
When attention, cache, and audio-side protection are combined, the full method achieves the best overall quality among the ablated variants.
These results show the improvement comes from explicitly protecting audio-video interaction, rather than merely increasing computation.

\paragraph{Protection ratio.}
Table~\ref{tab:ratio_ablation} evaluates the effect of the protection ratio.
A small ratio of $2\%$ keeps more sparse computation and gives higher speedup, but leaves more synchronization-critical queries unprotected.
Increasing the ratio to $10\%$ further improves generation quality and synchronization metrics, but the additional dense recomputation reduces the speedup.
We therefore use $5\%$ as the default setting, which provides a practical trade-off between generation quality, synchronization preservation, and inference efficiency.

\subsection{Qualitative Analysis}

Fig.~\ref{fig:qualitative_comparison} compares dense inference, unprotected acceleration, and our protected acceleration.
Without protection, accelerated inference is more likely to introduce artifacts, identity drift, or temporal inconsistency in salient regions.
In contrast, our method remains closer to the dense baseline by preserving high-fidelity computation for regions indicated by cross-modal saliency.
This qualitative result supports the importance of the proposed protection mechanism for keeping generation quality under sparse or cached computation.

\begin{table}[t]
\centering
\small
\setlength{\tabcolsep}{5pt}
\caption{
Component ablation under sparse computation.
Checkmarks indicate whether the corresponding protection strategy is enabled.
}
\label{tab:component_ablation}
\resizebox{\linewidth}{!}{%
\begin{tabular}{ccc|ccccc}
\toprule
Attn & Cache  & Audio  &
PSNR$\uparrow$ & SSIM$\uparrow$ & LPIPS$\downarrow$ & Sync-C$\uparrow$ & Speedup$\uparrow$\\
\midrule
           &            &            & 21.5338 & 0.7689 & 0.2462 & 3.626 & \textbf{2.230x} \\
\checkmark &            &            & 24.2551 & 0.8182 & 0.1533 & 3.939 & 2.152x \\
           & \checkmark &            & 22.9203 & 0.7821 & 0.1915 & 3.852 & 2.063x \\
\checkmark & \checkmark & \checkmark & \textbf{25.6857} & \textbf{0.8385} & \textbf{0.1398} & \textbf{4.606} & 1.992x \\
\bottomrule
\end{tabular}
}
\end{table}

\begin{table}[t]
\centering
\small
\setlength{\tabcolsep}{5pt}
\caption{
Ablation study on the protection ratio.
}
\label{tab:ratio_ablation}
\resizebox{\linewidth}{!}{%
\begin{tabular}{c|ccccc}
\toprule
Prot. Ratio & PSNR$\uparrow$ & SSIM$\uparrow$ & LPIPS$\downarrow$ & Sync-C$\uparrow$ & Speedup$\uparrow$ \\
\midrule
2\%  & 24.8926 & 0.8133 & 0.1672 & 4.539 & \textbf{2.02x} \\
5\%  & 25.6857 & 0.8385 & 0.1398 & 4.606 & 1.992x \\
10\% & \textbf{25.8201} & \textbf{0.8413} & \textbf{0.1369} & \textbf{4.648} & 1.861x \\
\bottomrule
\end{tabular}
}
\end{table}
\section{Conclusion}

We present an audio-guided protected sparse attention framework for efficient audio-visual generation.
The core idea is to use cross-modal attention as sound-source saliency and protect audio-critical video tokens during sparse video self-attention.
By aligning the acceleration policy with audio-video interaction, the method aims to reduce redundant computation while preserving the visual evidence needed for synchronized audio generation.

{
\small
\bibliographystyle{ieeenat_fullname}
\bibliography{references}
}

\clearpage
\appendix

\twocolumn[
\begin{center}
    {\Large \bf Supplementary Material}
\end{center}
\vspace{1em}
]

\section{Overview}
This supplementary material consists of:

\begin{itemize}
    \item Evaluation metrics ;
    \item Dual-stream backbone architecture ;
    \item Implementation details ;
    \item Saliency stability analysis ;
    \item Additional qualitative results .
\end{itemize}



\section{Evaluation Metrics}
\label{sec:evaluation_metrics}

Unless otherwise specified, full-reference video and audio metrics are computed by comparing the accelerated output with the dense baseline output generated with the same prompt, random seed, and sampling configuration.
For video-level metrics, we compute the score frame by frame and report the average over all frames and evaluation samples.

\paragraph{Peak Signal-to-Noise Ratio (PSNR).}
PSNR measures the pixel-level reconstruction fidelity between an accelerated video frame and its dense baseline reference.
Given a reference frame $I$ and an accelerated frame $\hat{I}$, PSNR is defined as
\begin{equation}
    \mathrm{PSNR}(I,\hat{I}) =
    10 \log_{10}\left(\frac{R^2}{\mathrm{MSE}(I,\hat{I})}\right),
\end{equation}
where $R$ is the maximum possible pixel value and $\mathrm{MSE}$ denotes the mean squared error between the two frames.
A higher PSNR indicates that the accelerated result is closer to the dense baseline in pixel space.

\paragraph{Structural Similarity Index Measure (SSIM)~\cite{wang2004image}.}
SSIM evaluates structural similarity by comparing luminance, contrast, and structural information between two images.
Given two frames $I$ and $\hat{I}$, SSIM is written as
\begin{equation}
    \mathrm{SSIM}(I,\hat{I}) =
    \frac{(2\mu_I\mu_{\hat{I}} + C_1)(2\sigma_{I\hat{I}} + C_2)}
    {(\mu_I^2+\mu_{\hat{I}}^2+C_1)(\sigma_I^2+\sigma_{\hat{I}}^2+C_2)},
\end{equation}
where $\mu$, $\sigma^2$, and $\sigma_{I\hat{I}}$ denote local means, variances, and covariance, respectively.
A higher SSIM indicates better structural fidelity.

\paragraph{Learned Perceptual Image Patch Similarity (LPIPS)~\cite{zhang2018unreasonable}.}
LPIPS measures perceptual distance using deep features extracted from a pretrained network.
Given feature extractor $\phi_l(\cdot)$ at layer $l$, LPIPS can be written as
\begin{equation}
    \mathrm{LPIPS}(I,\hat{I}) =
    \sum_l w_l \left\| \phi_l(I) - \phi_l(\hat{I}) \right\|_2^2,
\end{equation}
where $w_l$ denotes the learned layer-wise weighting.
A lower LPIPS indicates smaller perceptual degradation.

\paragraph{Q-Align~\cite{wu2023q}.}
Q-Align is a no-reference perceptual quality assessment metric built upon large multimodal models.
It formulates visual quality assessment as a level-alignment problem, where the model predicts quality scores by aligning visual content with discrete text-defined quality levels.
Unlike PSNR, SSIM, and LPIPS, Q-Align does not compare the accelerated video with a reference frame, but instead evaluates the perceptual quality of the generated visual content itself.
We use Q-Align as a complementary metric to measure whether acceleration preserves the overall perceptual quality of generated videos.
A higher Q-Align score indicates better perceptual visual quality.

\paragraph{Sync-C and Sync-D~\cite{chung2016out}.}
Sync-C and Sync-D are SyncNet-style metrics for measuring audio-video synchronization.
They embed the visual stream and the audio stream into a shared synchronization space and compare their temporal correspondence across possible offsets.
Sync-D denotes the distance between the best-matched audio-visual embeddings, where a lower value indicates a smaller synchronization error.
Sync-C measures the confidence of the best temporal alignment, reflecting how clearly the correct audio-video offset can be distinguished from other offsets.
A higher Sync-C and a lower Sync-D indicate better audio-video synchronization.

\paragraph{VISQOL~\cite{hines2015visqol}}
VISQOL is a full-reference objective audio quality metric designed to estimate perceptual similarity between a reference audio signal and a degraded audio signal.
It compares audio signals using perceptually motivated time-frequency representations and predicts an objective quality score correlated with human listening judgments.
In our evaluation, the dense baseline audio is used as the reference, and the accelerated audio is used as the evaluated signal.
Thus, VISQOL measures whether acceleration preserves the perceptual audio quality of the dense generation result.
A higher VISQOL score indicates better audio quality preservation.

\section{Dual-Stream Backbone Architecture}
\label{sec:dual_stream_backbone}

\begin{figure}[t]
\centering
\includegraphics[width=0.92\linewidth]{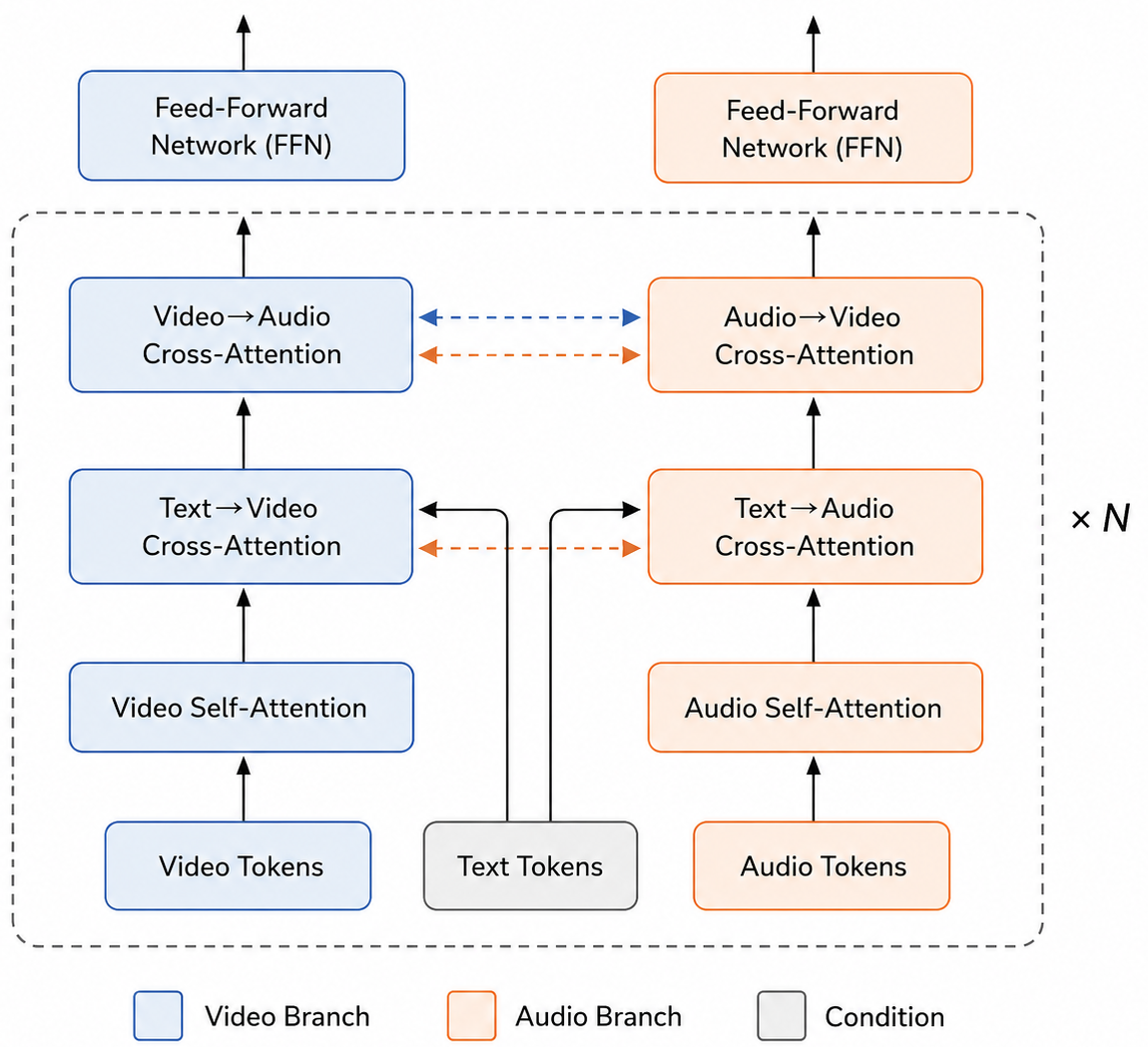}
\caption{
Illustration of a typical dual-stream audio-visual diffusion block.
The video and audio branches are processed separately and exchange information through bidirectional audio-video attention.
}
\label{fig:supp_dual_stream_backbone}
\end{figure}

This section provides a more detailed description of the dual-stream backbone used by recent joint audio-visual diffusion models such as Ovi~\cite{low2025ovi} and LTX-2~\cite{hacohen2026ltx}.
As shown in Fig.~\ref{fig:supp_dual_stream_backbone}, the model maintains two modality-specific token streams: a video stream for spatiotemporal visual latents and an audio stream for acoustic latents.
The two streams are processed by separate transformer branches, each containing self-attention and feed-forward layers.

Within a typical dual-stream block, modality-specific self-attention first updates video and audio tokens within their own representation spaces.
Text tokens are then used as semantic conditions through text-to-video and text-to-audio attention.
After that, bidirectional audio-video attention allows two modalities to exchange information: the video branch can attend to audio tokens, and the audio branch can attend to video tokens.
Stacking such blocks over denoising steps enables the model to jointly refine visual and acoustic content.

Our method operates on this backbone without changing its parameters or training procedure.
The cross-modal attention maps produced by the bidirectional audio-video attention modules are used only as saliency signals for identifying tokens or blocks that should be protected during acceleration.

\section{Implementation Details}
\label{sec:implementation_details}

We provide the main acceleration hyperparameters used in our experiments.
For protected sparse attention, we use a protection ratio of $5\%$ by default, meaning that only the top $5\%$ cross-modal salient queries are recomputed with dense attention rows.
The remaining queries follow the sparse attention path.
For vanilla sparse baselines, we use the same sparse attention budget but remove the protection mechanism.

For lagged cross-modal saliency, the saliency cache is updated every 4 denoising steps and reused for the nearby steps.
For cache-based acceleration, we keep full computation for 15\% of the denoising steps as anchor steps for refreshing cached features and saliency signals.
At the block level, 2 fusion blocks are kept under full computation, while cache reuse is applied to the remaining eligible blocks.

For synchronization-aware cache reuse, the protected-region weight is set to $\lambda_p=0.5$, and the saliency-drift weight is set to $\lambda_s=0.1$.
The cache reuse threshold is set to $\tau=0.2$.
A block output is reused only when the synchronization-aware indicator is below this threshold.
For LTX-2.3, acceleration is applied only to the low-resolution denoising stage, while the upsampling and high-resolution refinement stages are kept unchanged.

\section{Additional Ablation on Saliency Selection}
\label{sec:supp_saliency_selection}

To further verify whether the protected regions are effectively selected by cross-modal guidance, we compare different strategies for choosing the protected queries under the same sparse computation setting.
The no-protection baseline applies sparse computation without recomputing any query rows.
The random baseline protects the same ratio of queries but selects them randomly.
The video self-attention baseline selects protected queries according to intra-modal attention scores from the video branch.
Our final strategy selects the top $5\%$ queries according to cross-modal saliency.

As shown in Table~\ref{tab:supp_saliency_selection}, cross-modal saliency provides a more effective protection signal than modality-agnostic selection.
Compared with the no-protection baseline, it substantially improves video reconstruction quality and audio-video synchronization, indicating that the gain comes from protecting synchronization-related computation rather than simply adding more dense computation.

\begin{table}[t]
\centering
\small
\setlength{\tabcolsep}{5pt}
\caption{
Ablation on saliency selection strategies.
}
\label{tab:supp_saliency_selection}
\resizebox{\linewidth}{!}{%
\begin{tabular}{l|ccccc}
\toprule
Saliency Strategy &
PSNR$\uparrow$ & SSIM$\uparrow$ & LPIPS$\downarrow$ &
Sync-C$\uparrow$ & Sync-D$\downarrow$ \\
\midrule
No Protection              & 21.5338 & 0.7689 & 0.2462 & 3.626 & 8.653 \\
Random 5\%                 &  22.0219  & 0.7762  & 0.2376 & 3.819 & 8.578 \\
Video Self-Attn Score 5\%  &  23.5128  & 0.8025  & 0.2140  & 3.764 & 8.602 \\
Cross-Modal Saliency 5\%   & \textbf{25.6857} & \textbf{0.8385} & \textbf{0.1398} & \textbf{4.606} & \textbf{8.378} \\
\bottomrule
\end{tabular}
}
\end{table}

\section{Saliency Stability Across Denoising Steps}
\label{sec:saliency_stability}

The main paper uses a lagged cross-modal saliency cache to guide protected sparse attention, instead of recomputing the saliency map at every denoising step.
Here we provide an additional analysis to support this design.
The analysis is conducted on Ovi using the 15-th fusion block across all denoising steps .
For two denoising steps $t$ and $t+\Delta$, we compute the overlap between their protected token sets:
\begin{equation}
    \mathrm{Overlap}(t,t+\Delta)
    =
    \frac{|\mathcal{P}_t \cap \mathcal{P}_{t+\Delta}|}{|\mathcal{P}_t|},
\end{equation}
where $\mathcal{P}_t$ denotes the top protected tokens selected from cross-modal saliency at step $t$.
A higher overlap means that the saliency-derived protected regions are more consistent across the two denoising steps.

\begin{figure}[t]
\centering
\includegraphics[width=0.85\linewidth]{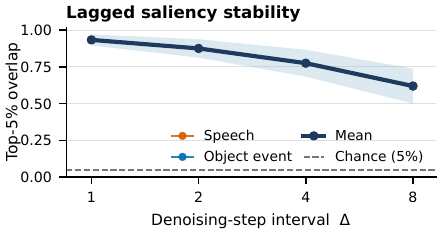}
\caption{
Protected-token overlap across denoising-step intervals.
The high overlap at small intervals supports lagged saliency reuse, while the decrease at larger intervals motivates periodic saliency updates.
}
\label{fig:supp_token_overlap}
\end{figure}

As shown in Fig.~\ref{fig:supp_token_overlap}, adjacent denoising steps have a high overlap of about $0.93$, and the overlap remains around $0.88$ at $\Delta=2$.
When the interval increases, the overlap gradually decreases, reaching about $0.78$ at $\Delta=4$ and about $0.62$ at $\Delta=8$.
This indicates that cross-modal saliency is stable over nearby steps but still evolves over longer intervals.
Therefore, the saliency map can be safely reused within a short temporal window to reduce overhead, while periodic updates remain necessary to track the changing audio-video interaction.
We also report the effect of different saliency update intervals in Table~\ref{tab:supp_saliency_update_interval}.
Updating saliency more frequently with $K=2$ brings only marginal improvement over $K=4$, while requiring more frequent cross-modal saliency recomputation.
In contrast, using a larger interval such as $K=8$ leads to a clear quality drop, indicating that the cached saliency becomes less reliable when reused for too many denoising steps.
Therefore, we use $K=4$ as the default setting to balance saliency accuracy and computational overhead.
\begin{table}[t]
\centering
\small
\setlength{\tabcolsep}{5pt}
\caption{
Ablation on the saliency update interval.
}
\label{tab:supp_saliency_update_interval}
\resizebox{\linewidth}{!}{%
\begin{tabular}{c|ccccc}
\toprule
$K$ &
PSNR$\uparrow$ & SSIM$\uparrow$ & LPIPS$\downarrow$ &
Sync-C$\uparrow$ & Sync-D$\downarrow$ \\
\midrule
2 & \textbf{25.7033} & \textbf{0.8408} & \textbf{0.1378} & \textbf{4.629} & 8.492 \\
4 & 25.6857 & 0.8385 & 0.1398 & 4.606 & \textbf{8.378} \\
8 & 24.3362 & 0.8166 & 0.1702 & 4.325 & 8.522 \\
\bottomrule
\end{tabular}
}
\end{table}

\section{Additional Qualitative Results}
\label{sec:additional_qualitative_results}

\begin{figure*}[t]
\centering
\includegraphics[width=0.95\textwidth]{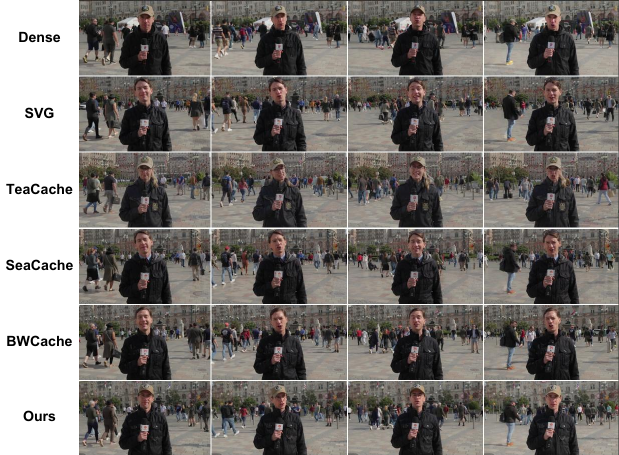}

\vspace{2mm}

\includegraphics[width=0.95\textwidth]{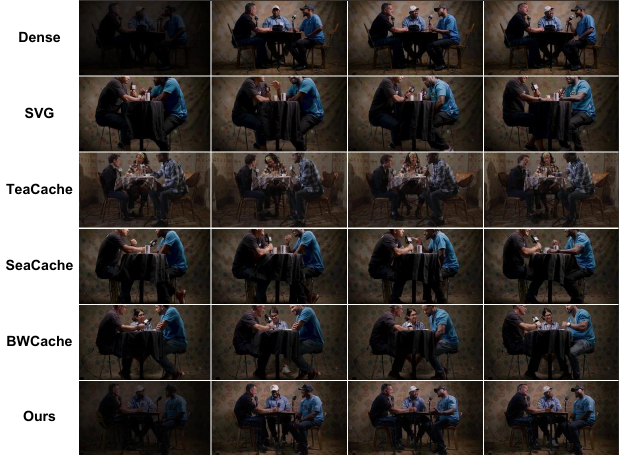}
\caption{
Additional qualitative comparisons on LTX-2.3.
}
\label{fig:supp_qual_1}
\end{figure*}

\begin{figure*}[t]
\centering
\includegraphics[width=0.95\textwidth]{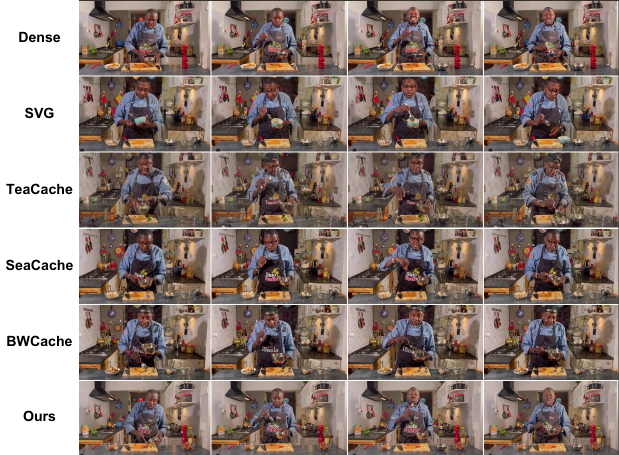}

\vspace{2mm}

\includegraphics[width=0.95\textwidth]{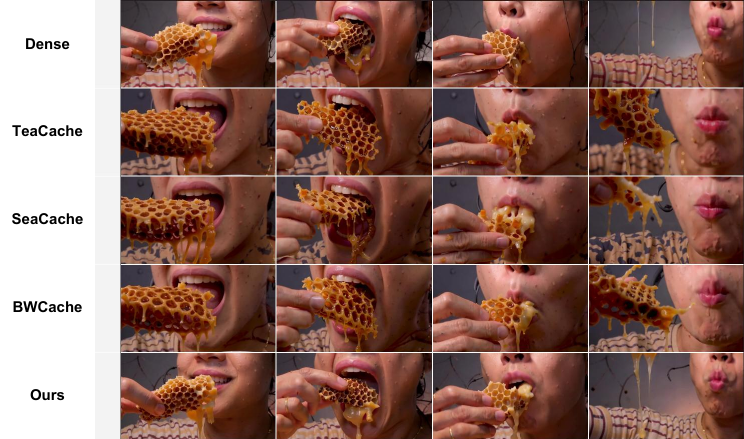}
\caption{
Additional qualitative comparisons on LTX-2.3.
}
\label{fig:supp_qual_2}
\end{figure*}

\begin{figure*}[t]
\centering
\includegraphics[width=0.95\textwidth]{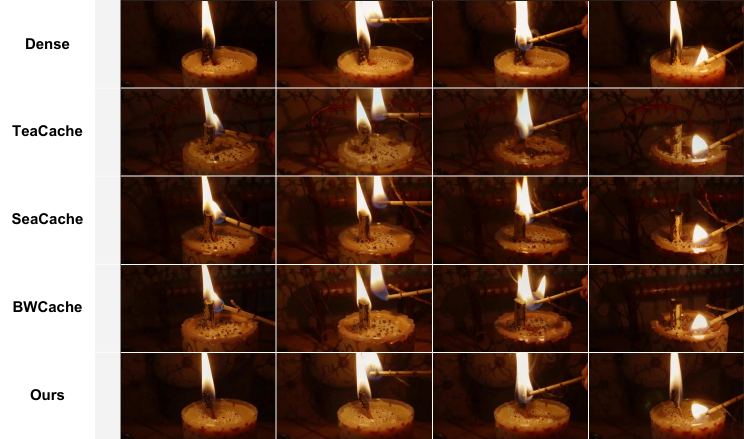}

\vspace{2mm}

\includegraphics[width=0.95\textwidth]{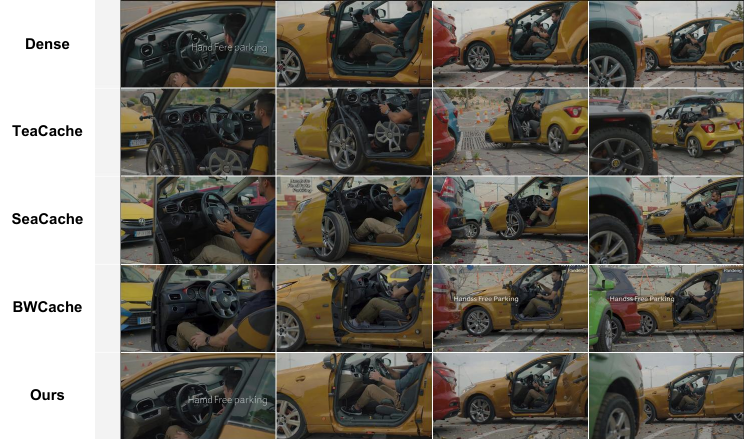}
\caption{
Additional qualitative comparisons on LTX-2.3.
}
\label{fig:supp_qual_3}
\end{figure*}

\begin{figure*}[t]
\centering
\includegraphics[width=0.95\textwidth]{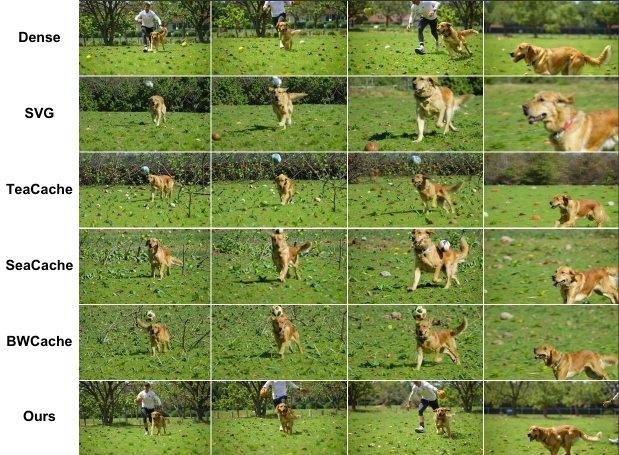}

\vspace{2mm}

\includegraphics[width=0.95\textwidth]{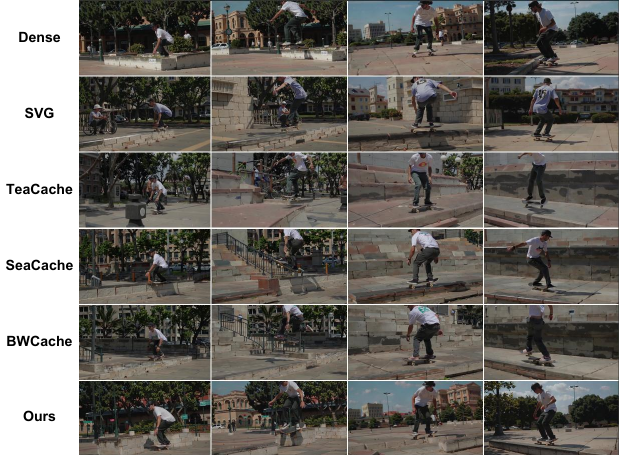}
\caption{
Additional qualitative comparisons on LTX-2.3.
}
\label{fig:supp_qual_4}
\end{figure*}

We provide additional qualitative comparisons on LTX-2.3 in Figs.~\ref{fig:supp_qual_1}--\ref{fig:supp_qual_4}.
Each example compares dense inference, existing acceleration baselines, and our method using four representative frames sampled from the generated video.
These visualizations complement the quantitative results by showing how different acceleration strategies affect local visual fidelity and temporal consistency.
Across diverse scenarios, including dialogue, object interaction, ASMR-style close-up motion, and animal or human motion, unprotected sparse or cache-based acceleration can introduce local distortions, inconsistent structures, or visible deviations from the dense baseline.
In contrast, our method better preserves salient foreground regions and key motion details while still reducing inference cost.
These qualitative results further support the effectiveness of synchronization-aware protection for maintaining generation quality under acceleration.

\end{document}